\documentclass{ceurart}

\usepackage{listings}
\usepackage{subcaption}
\usepackage{svg}
\usepackage{url}

\begin{document}

\copyrightyear{2026}
\copyrightclause{Copyright for this paper by its authors.
  Use permitted under Creative Commons License Attribution 4.0
  International (CC BY 4.0).}

\conference{EVALITA 2026: 9th Evaluation Campaign of Natural Language
Processing and Speech Tools for Italian, Feb 26 – 27, Bari, IT}

\title{KIT-TIP-NLP at MultiPride: Continual Learning with Multilingual Foundation Model}

\tnotemark[1]
\tnotetext[1]{This paper contains examples of explicitly offensive content.}

\author[]{Barathi Ganesh HB}[%
orcid=0000-0002-1150-2773,
email=hbbg.jp@gmail.com,
]
\address{Text Information Processing Lab, Kitami Institute of Technology, Kitami, Hokkaido 090-0015, Japan}

\author[]{Michal Ptaszynski}[%
orcid=0000-0002-1910-9183,
email=michal@mail.kitami-it.ac.jp,
]

\author[]{Rene Melendez}[%
orcid=0009-0004-2129-8747
]

\author[]{Juuso Eronen}[%
orcid=0000-0001-9841-3652
]

\begin{abstract}
This paper presents a multi-stage framework for detecting reclaimed slurs in multilingual social media discourse. It addresses the challenge of identifying reclamatory versus non-reclamatory usage of LGBTQ+-related slurs across English, Spanish, and Italian tweets. The framework handles three intertwined methodological challenges like data scarcity, class imbalance, and cross-linguistic variation in sentiment expression. It integrates data-driven model selection via cross-validation, semantic-preserving augmentation through back-translation, inductive transfer learning with dynamic epoch-level undersampling, and domain-specific knowledge injection via masked language modeling. Eight multilingual embedding models were evaluated systematically, with XLM-RoBERTa selected as the foundation model based on macro-averaged F1 score. Data augmentation via GPT-4o-mini back-translation to alternate languages effectively tripled the training corpus while preserving semantic content and class distribution ratios. The framework produces four final runs for the evaluation purposes where RUN 1 is inductive transfer learning with augmentation and undersampling, RUN 2 with masked language modeling pre-training, RUN 3 and RUN 4 are previous predictions refined via language-specific decision thresholds optimized via ROC analysis. Language-specific threshold refinement reveals that optimal decision boundaries vary significantly across languages. This reflects distributional differences in model confidence scores and linguistic variation in reclamatory language usage. The threshold-based optimization yields 2–5\% absolute F1 improvement without requiring model retraining. The methodology is fully reproducible, with all code and experimental setup are available at \url{https://github.com/rbg-research/MultiPRIDE-Evalita-2026}.
\end{abstract}

%%
%% Keywords. The author(s) should pick words that accurately describe
%% the work being presented. Separate the keywords with commas.
\begin{keywords}
Reclamation Detection \sep
  Multilingual Foundation Models \sep
  Transfer Learning \sep
  Dynamic Undersampling \sep
  Back-Translation Augmentation \sep
  Language-Specific Decision Boundaries \sep
  LGBTQ+ Sentiment Analysis
\end{keywords}

%%
%% This command processes the author and affiliation and title
%% information and builds the first part of the formatted document.

\maketitle

\section{Introduction}

Automated content moderation systems frequently struggle to distinguish between hate speech and reclaimed language \cite{zsisku2024hate, chakravarthi2022overview}. Reclaimed slurs are historically offensive terms that marginalized communities, particularly LGBTQ+ groups which reappropriated for self-empowerment and solidarity \cite{popa2020reclamation, evalita2026overview}. When algorithms fail to recognize this context, they often flag harmless posts as toxic or miss actual hate speech. This paper describes the KIT-TIP-NLP system developed for the MultiPride task at EVALITA 2026, which focuses on detecting reclaimed slurs in English, Spanish, and Italian tweets \cite{evalita2026overview}.

The task presents three main obstacles. Firstly, the available data is limited. Secondly, the datasets are highly imbalanced, with far fewer examples of reclaimed usage than non reclaimed usage. Thirdly, the way speakers reclaim slurs varies by language and culture. A model that works well for English sarcasm may fail to grasp Italian cultural markers.

In order to tackle these problems, we employed a multistage framework. The first step was to assess a variety of multilingual models and proceed with XLM-RoBERTa as our basis. To tackle the problem of insufficient data, we implemented the GPT-4o-mini-based back-translation method that resulted in the training set being tripled while the original meaning was preserved. Additionally, we made use of a proactive sampling method during the training to ensure that the model does not turn a blind eye to the minority class. To improve performance further, we assigned language-specific decision thresholds instead of applying one universal cutoff for all languages. This method assumes that the model has a higher level of confidence in certain languages and hence provides a way to get better accuracy without cumbersome retraining.

The rest of the paper is structured in the following way. Section \ref{sec:method} describes the methodology with a focus on data analysis, model selection, and the training process. Section \ref{sec:experiments} describes the experimental setup and the performance of the four runs. Section \ref{sec:observations} discusses the results and the challenges faced in the different languages. Finally, section \ref{sec:conclusion} summarizes the findings.

\section{Methodology}
\label{sec:method}

\begin{figure}[!ht]
  \centering
  \includegraphics[width=\linewidth-80pt]{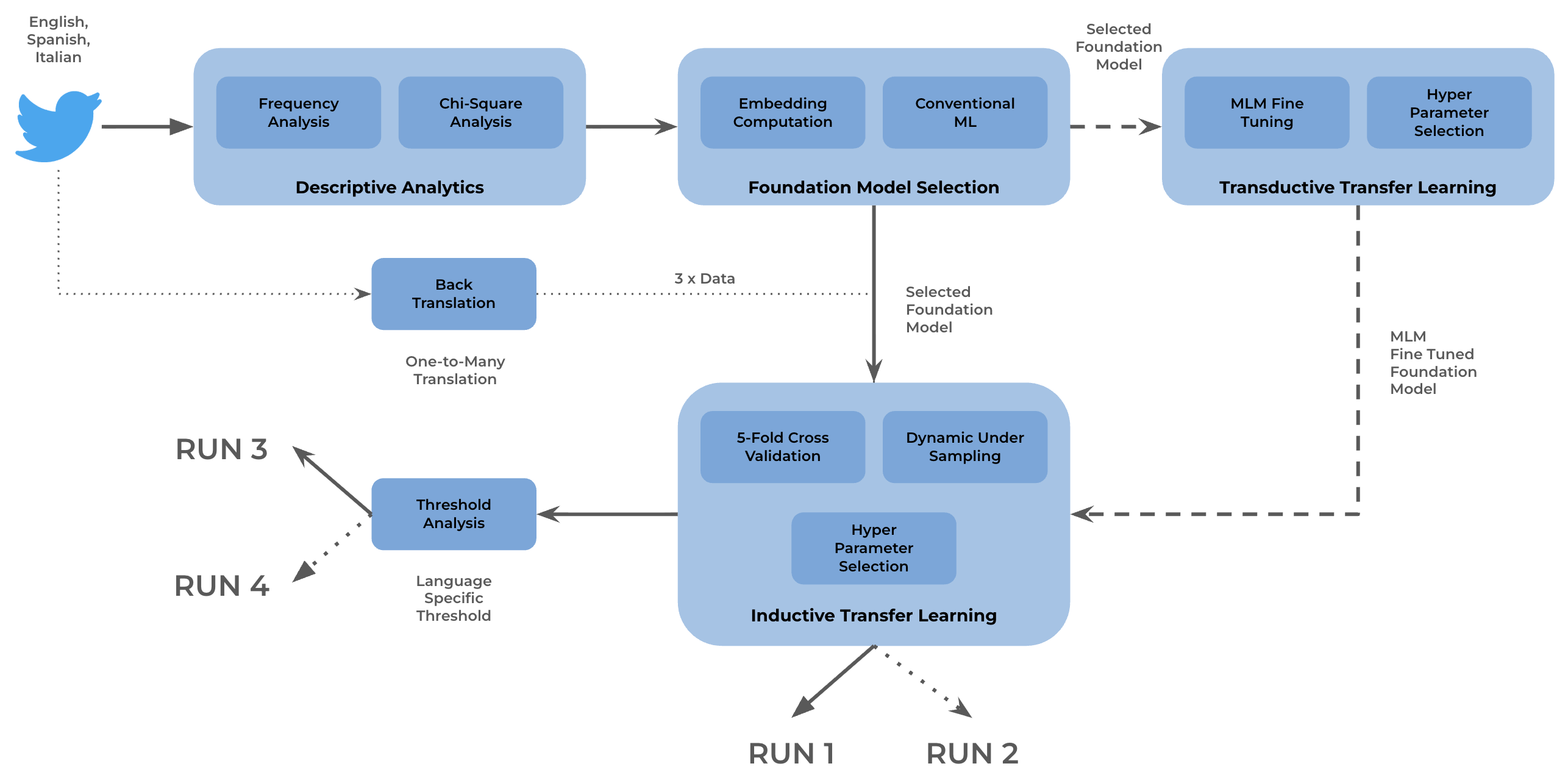}
  \caption{Multi-stage multilingual hate-speech classification framework with four sequential runs refining performance via data-driven model selection, augmentation, hyperparameter selection, 5-fold CV, MLM adaptation, and threshold calibration. \textbf{RUN 1:} Inductive transfer learning with optimal foundation model. \textbf{RUN 2:} Transductive transfer learning on optimal foundation model follwed by Inductive transfer learning. \textbf{RUN 3:} Threshold refinement on RUN 1 outputs and \textbf{RUN 4:} threshold refinement on RUN 2 outputs to handle linguistic nuances and cross-lingual calibration.}
  \label{arch}
\end{figure}

% The overall framework, as illustrated in Figure \ref{arch}, proceeds through four sequential research runs that progressively refine classification performance. This multi-layered approach reflects four core design principles. First, the framework prioritizes data-driven model selection through systematic cross-validation, mitigating selection bias and ensuring that foundation model choice is evidenced with empirical performance. Second, it addresses data scarcity through semantic-preserving augmentation, expanding the effective training corpus while maintaining label distribution through back-translation. Third, the integration of domain-specific knowledge via masked language modeling (MLM) for optimal representation. Finally language-specific threshold refinement for handling linguistic nuance and distributional variation across languages. The progression from RUN 1 through RUN 4 allows for systematic ablation and comparison of these design choices.

The overall framework, as represented in Figure \ref{arch}, consists of four research runs carried out one after another which are able to refine classification performance in a step-wise manner. The multi-layered approach goes back to four design principles. First, the framework follows the data-driven model selection through systematic cross-validation as its primary consideration which eliminates the possibility of selection bias and guarantees that the chosen foundation model comes with empirical performance. Second, it is based on the scarcity of data that is through semantic-preserving augmentation, thus increasing the effective training corpus while keeping the distribution of labels the same using the method of back-translation. Thirdly, the domain-specific knowledge is integrated by masked language modeling (MLM) for the best representation. Finally, language-specific threshold refinement is done for managing the linguistic nuance and the distributional variation in different languages. The whole process from RUN 1 to RUN 4 provides the opportunity for systematic ablation and comparison of these design choices.

\subsection{Data Analysis and Bias Detection}
% The framework starts with a descriptive analytics step to characterize the dataset\footnote{https://multipride-evalita.github.io/, Accessed on January 2026.} and identify potential imbalances with respect to language and labels. Frequency analysis was performed across all labels and languages to examine the distribution of positive and negative instances. Chi-square analysis was subsequently applied to detect statistical associations between language and label distributions \cite{tallarida1987chi}. This reveals inherent label and language biases present in the provided dataset for subsequent modeling decisions.

The framework commences by performing descriptive analytics on the dataset \footnote{https://multipride-evalita.github.io/, Accessed on January 2026.} to characterize it and recognize possible imbalances regarding language and labels. A frequency distribution analysis was done for all labels and languages to check the positive and negative instance distribution. After that, chi-square analysis was conducted to discover the statistical relationships between language and label distributions \cite{tallarida1987chi}. This exposes the label and language biases naturally existing in the dataset supplied for the modeling choices made afterwards.

\subsection{Foundation Model Selection}
This phase employed a two-stage approach. First, embedding representations were computed for all the tweets with each foundation models for assessing the capability of representating semantic and linguistic properties of the input tweet. A conventional machine learning (ML) pipeline was established in parallel to provide a comparative reference. Five-fold cross-validation was conducted across both embedding-based and conventional ML approaches to ensure robust and unbiased model selection. The cross-validation framework evaluated across multiple models and selected the optimal foundation model based on performance metrics computed across all folds for avoiding the selection bias.

\subsection{Inductive Transfer Learning with Data Augmentation}
\label{sec:itl}

The chosen foundational model went through an inductive transfer learning approach \cite{pan2010survey} for fine-tuning. A back-translation augmentation method was executed to deal with data deficiency with GPT-4o-mini being used as the translation engine \cite{openai2024gpt4omini}. Thru the supplied dataset, each tweet in the original language was assessed and translated to the other two languages in a systematic way. This allows for the generation of semantically equal, yet syntactically diverse paraphrases. This one-to-many translation operation has effectively increased the size of the training set threefold. The augmented dataset was then used to refine the base model further \cite{taheri2025enhancing}.

In order to address the issue of class imbalance in the classification task, a dynamic under-sampling strategy was used at the epoch level \cite{pouyanfar2018dynamic}. The sampling technique during each training epoch allowed every positive class instance in the training batch to have three negative class instances selected randomly without replacement. This 1:3 ratio was dynamically sustained across all epochs enabling the model to learn distinguishing features without being affected by class imbalance. At Each epoch this sampling technique was random which results in stochasticity that enhanced the learning generalization. Five-fold cross-validation was applied not only for the verification of the final model accuracy but also the hyper-parameter tuning. This final result out of this phase is marked as RUN 1.

\subsection{Domain Knowledge Integration via Masked Language Modeling}
% In a second experimental pathway, MLM was performed on the foundation model to inject domain-specific linguistic knowledge. The MLM pretraining task involves predicting randomly masked tokens in sequences, helps the model develop deeper linguistic representations specific to the target domain. Following this adaptation, hyperparameter selection was performed using Optuna, an automated hyperparameter optimization framework that employs efficient sampling and pruning algorithms. The final MLM adapted model was then subjected to the same inductive fine-tuning steps as described in the previous section that includes back-translation augmentation, dynamic epoch level under-sampling at the 1:3 positive-to-negative ratio, and five-fold cross-validation for fitness assessment. This outcome result is designated as RUN 1.

In a second experimental pathway, masked language modeling (MLM) was applied to the foundation model for the purpose of inserting linguistic knowledge specific to the domain. The MLM pretraining task involves predicting randomly masked tokens in sequences, helps the model develop deeper linguistic representations specific to the target domain. The hyperparameter selection was then done with the help of Optuna, which is an automated hyperparameter optimization tool that employs efficient sampling and pruning algorithms. The final model that had undergone MLM adaptation was again put through the same process of inductive fine-tuning as stated earlier in the section \ref{sec:itl}, which included back-translation augmentation, dynamic epoch-level under-sampling at the 1:3 positive-to-negative ratio, and five-fold cross-validation for fitness assessment. This final result out of this phase is marked as RUN 2.

\subsection{Language-Specific Threshold Refinement}
To further improve classification performance, language specific decision boundaries were derived for each language in the dataset. Confidence scores from the classifier were analyzed separately for English, Spanish, and Italian tweets. Language-specific thresholds were determined that optimally balanced precision and recall for each language through the ground truth labels. This step helps in accounting the distributional differences in the model's prediction scores. These thresholds were finally applied to reclassify predictions from both RUN 1 and RUN 2, resulting RUN 3 and RUN 4 respectively. This refinement recognizes that optimal decision thresholds may differ across languages due to linguistic variation and also different patterns of language usage in the context of reclaimed slurs.

\subsection{Evaluation Metrics and Statistical Analysis}
All results were evaluated using standard metrics appropriate for imbalanced binary classification. Reported metrics includes averaging F1 across classes equally for minimizing majority class dominance (Macro-averaged F1 score), Precision and Recall with per class and language breakdowns, and Area Under the ROC Curve for threshold-independent measure (ROC-AUC).

Model selection during all stages of the experimentation mainly relied on macro-averaged F1 calculated through stratified 5-fold cross-validation. With this approach, the minority (reclamatory) class is treated equally together with the majority class. It is a reasonable step to take in the context of the shared task, in which both false positives and false negatives have semantic importance. Confidence intervals (95\%) on cross-validation scores were computed using a set of samples to quantify the uncertainty in the performance estimates.

The language-specific performance breakdowns were calculated to evaluate multilingual generalization and pinpoint challenges. The optimization histories of Optuna were visualized using parameter importance plots and trial convergence curves to know which hyperparameters had the most impact on the validation F1 score and how rapidly the TPE sampler reached the high-quality solutions.

% Model selection across all experimental phases prioritized macro-averaged F1 computed via stratified 5-fold cross-validation. This criterion ensures that the minority (reclamatory) class receives equal consideration with the majority class. This step is appropriate for the shared task context where both false positives and false negatives carry semantic significance. Confidence intervals (95\%) on cross-validation scores were computed using set of samples to quantify uncertainty in performance estimates.

% Language specific performance breakdowns were computed to assess multilingual generalization and identify challenges. Optuna optimization histories were visualized via parameter importance plots and trial convergence curves to understand which hyper parameters most influenced validation F1 and how quickly the TPE sampler converged to the high quality solutions. 

\section{Experiments and Results}
\label{sec:experiments}

% All experiments were conducted on RTX4090 GPU clusters with CUDA 12.6 and cuDNN 8.9 support. Training leveraged automatic mixed precision (AMP) with bfloat16 for computational efficiency while maintaining gradient stability. Experiment tracking was performed via logging hyperparameters, metrics, loss curves, and model artifacts across all experimental runs. All code was implemented in Python 3.10 using standard scientific libraries.

All the experiments were carried out on RTX4090 GPU workstation having the support of CUDA 12.6 and cuDNN 8.9. The training employed automatic mixed precision (AMP) turning to bfloat16 for the purpose of computational efficiency and at the same time keeping the gradients stable. Experiment tracking was accomplished through the logging of hyperparameters, metrics, loss curves, and model artifacts for all experimental runs. All the code was done in Python 3.10 making use of the standard scientific libraries.

\begin{figure}[!ht]
  \centering
  \begin{subfigure}[b]{0.48\textwidth}
    \centering
    \includegraphics[width=\textwidth]{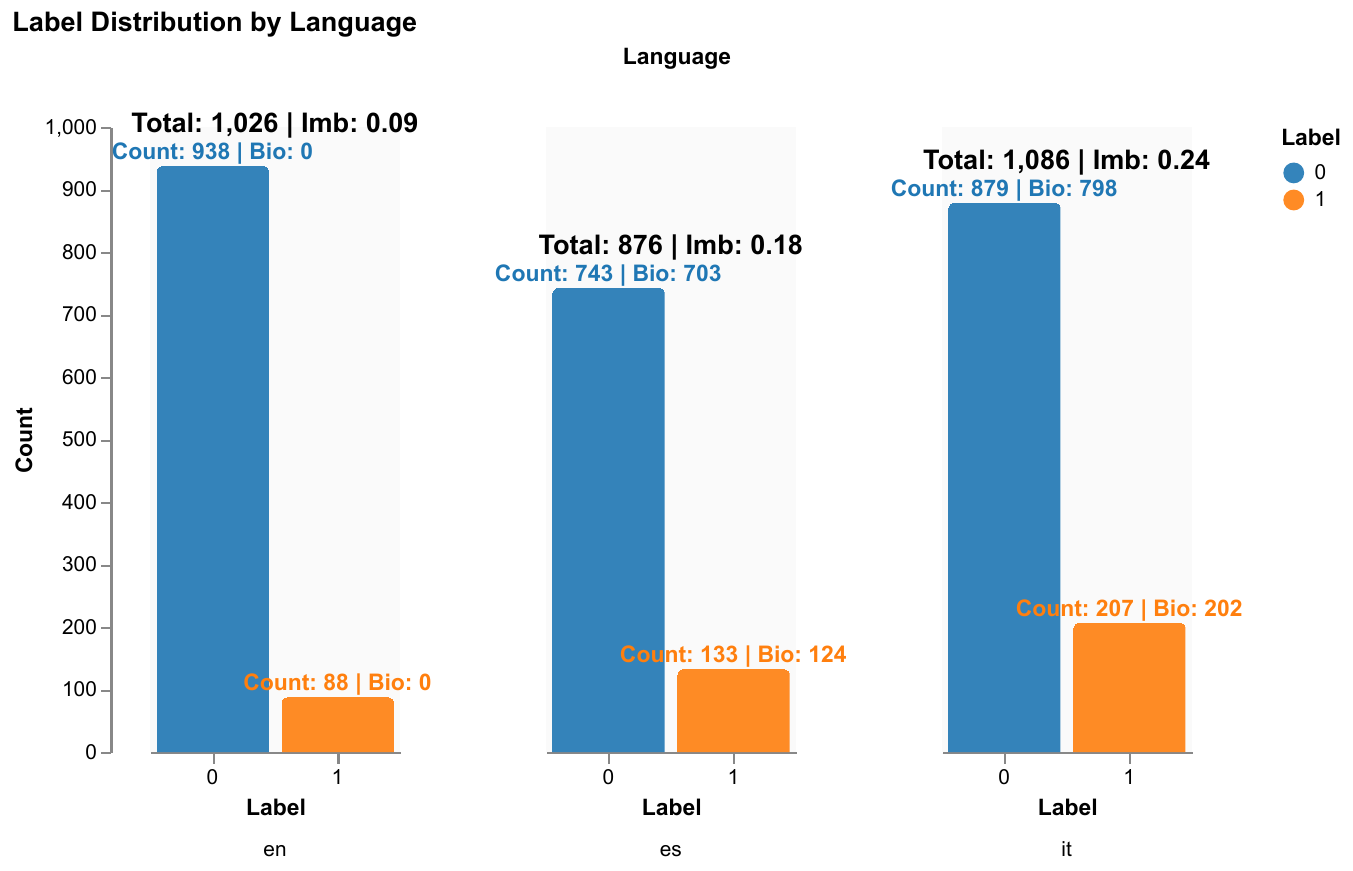}
    \caption{Label Distribution - Raw Data}
    \label{fig:image1}
  \end{subfigure}
  \hfill
  \begin{subfigure}[b]{0.48\textwidth}
    \centering
    \includegraphics[width=\textwidth]{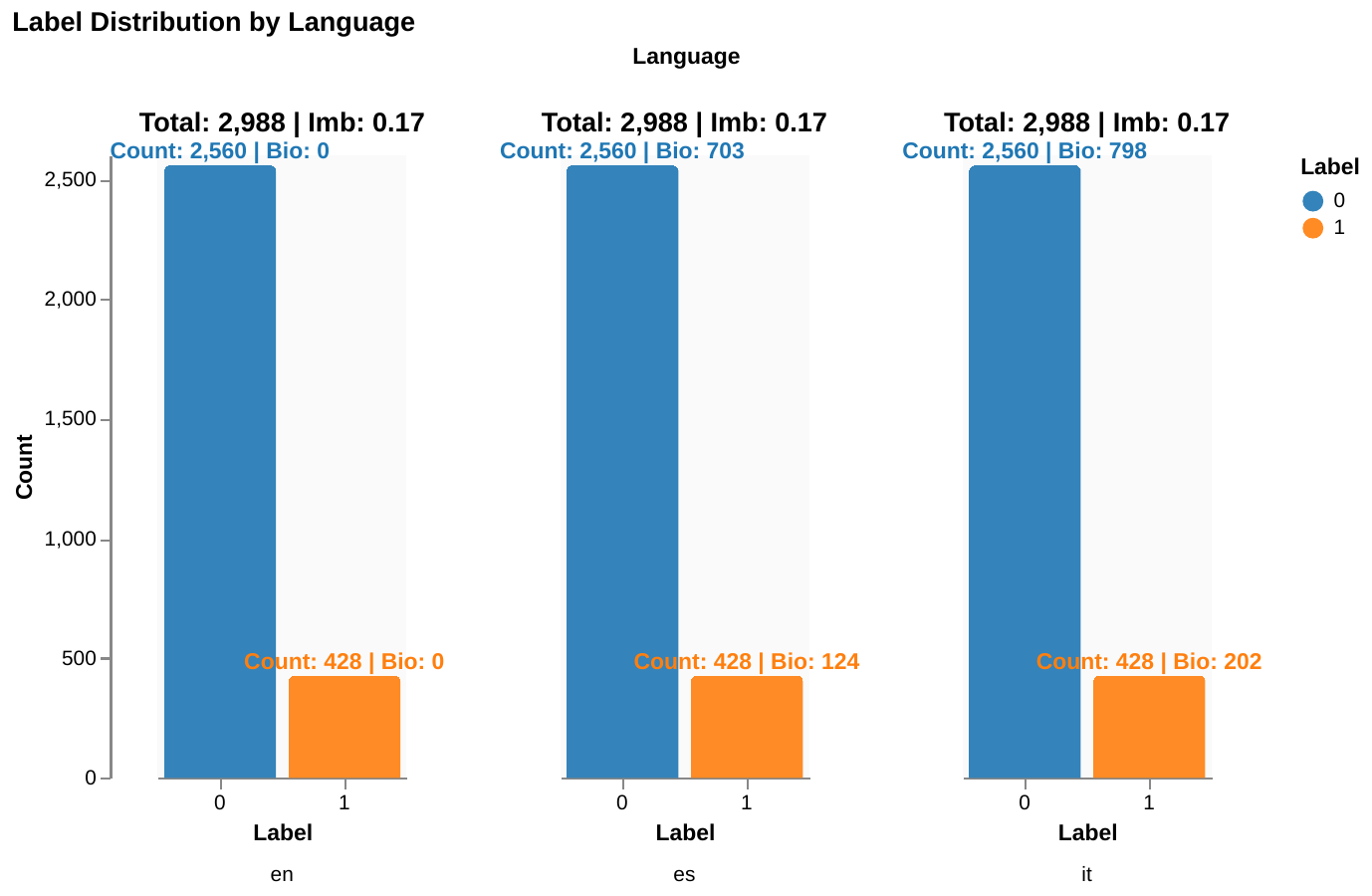}
    \caption{Label Distribution - Augumented Data}
    \label{fig:image2}
  \end{subfigure}

  \vspace{0.5cm}

  \begin{subfigure}[b]{0.80\textwidth}
    \centering
    \includegraphics[width=\textwidth]{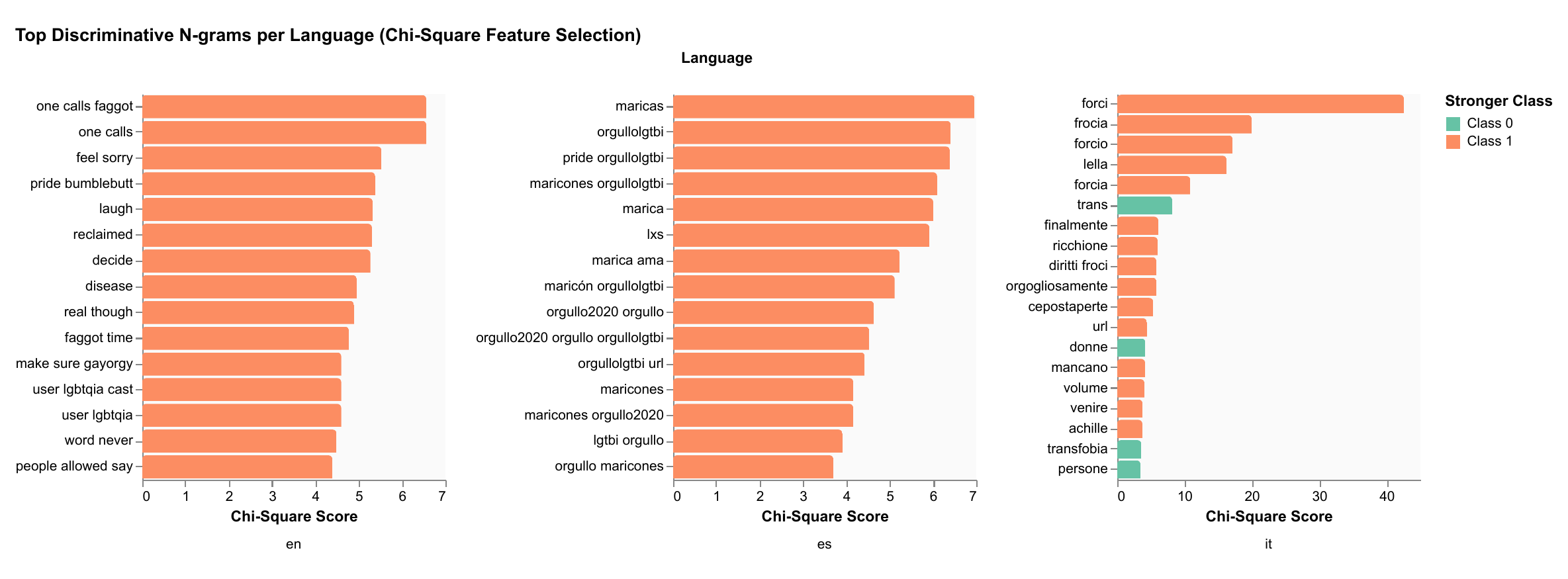}
    \caption{Chi-square Analysis}
    \label{fig:image3}
  \end{subfigure}

  \caption{Data distribution statistics: label imbalance in original dataset, augmented dataset after back-translation, and chi-square analysis of language-label associations.}
  \label{data_statistics}
\end{figure}

\subsection{Data Characteristics and Descriptive Analytics}
% The MultiPRIDE dataset comprises tweets in English, Spanish, and Italian labeled for reclamatory versus non-reclamatory usage of slurs in LGBTQ+ contexts. Initial exploratory analysis revealed significant class imbalance and language-specific label distribution heterogeneity. Frequency analysis across the full dataset demonstrated that the reclamatory class represented approximately 9-24\% of samples, with per-language variance: English (9\%), Spanish (18\%), Italian (24\%). Chi-square tests \cite{tallarida1987chi} of independence between language and label yielded $\chi^{2}$ values exceeding critical thresholds (p < 0.001), indicating non random label distribution across languages. 

The MultiPRIDE dataset contains slurs in LGBTQ+ contexts labeled as reclamatory or non-reclamatory for usage through tweets in English, Spanish, and Italian. The first exploratory analysis uncovered a major imbalance between the classes and a distribution of labels that varied by language. The full dataset frequency analysis revealed that the reclamatory class accounted for about 9-24\% of the samples with a different distribution per language: English (9\%), Spanish (18\%), and Italian (24\%). Chi-square tests \cite{tallarida1987chi} of independence between language and label produced $\chi^{2}$ values that were above the critical thresholds (p < 0.001), which meant that the labels were not randomly distributed over the languages. This finding justified the subsequent implementation of language specific decision thresholds and stratified cross validation splitting to ensure fold wise class distribution consistency. The data statistics are depicted in Figure \ref{data_statistics}.

\subsection{Foundation Model Selection and Baseline Evaluation}
Eight multilingual embedding models were evaluated as candidates for downstream fine-tuning: multilingual-e5-large (state-of-the-art dense retrieval embeddings with 1024 dimensions) \cite{wang2024multilingual}, bge-m3 (BGE multilingual model optimized for semantic search) \cite{chen2024bge}, gte-multilingual-base (General-purpose Text Embeddings with 768 dimensions) \cite{zhang2024mgte}, jina-embeddings-v3 (Jina's multilingual dense representation model) \cite{sturua2024jina}, snowflake-arctic-embed-l-v2.0 (Snowflake's large-scale multilingual embeddings) \cite{snowflake2024arctic}, LaBSE (Language-agnostic BERT Sentence Embeddings, 768 dimensions) \cite{feng2022language}, USE-multilingual (Universal Sentence Encoder for 16+ languages) \cite{yang2020multilingual}, and XLM-RoBERTa-large (550M parameters, 24 layers, enhanced capacity for cross-lingual transfer) \cite{conneau2020unsupervised}. For each candidate model, dense representations were computed by extracting contextualized embeddings from the final transformer layer, yielding 768-dimensional or 1024-dimensional vectors depending on the model architecture. These embeddings were subsequently used as input features for classifier training in the foundation model selection phase.

% A stratified 5-fold cross validation framework was instantiated, ensuring class distribution preservation across folds (80\% training, 20\% validation per fold). Within each fold, conventional ML baselines were trained on the computed embeddings with linear Support Vector Machine (Linear SVC with C=1.0, dual=False for efficiency). Evaluation metrics computed across all folds included macro-averaged Accuracy, Precision, Recall, and F1 score. Model selection prioritized on macro-averaged F1 stability (low fold-wise variance) over maximum absolute performance, as this indicates robustness to data distribution shifts and generalization capability. Analysis from Table \ref{tab:svc_balanced} revealed that XLM-RoBERTa-large achieved optimal balance between performance (macro F1 = 0.76 ± 0.04 across folds) and computational efficiency, justifying its selection as the foundation model for all downstream experiments.

A stratified 5-fold cross-validation framework was set up, which kept the same distribution of classes in the folds (80\% training, 20\% validation for each fold). Conventional machine learning baselines were then trained using the computed embeddings within each fold, applying linear Support Vector Machine (Linear SVC with C=1.0, dual=False for the sake of speed). The evaluation metrics computed for all folds include macro-averaged Accuracy, Precision, Recall, and F1 score. Besides, model selection preferred the macro-averaged F1 stability (low fold-wise variance) over the maximum absolute performance which means that the model is robust to data distribution shifts and can generalize. The analysis in Table \ref{tab:svc_balanced} showed that XLM-RoBERTa-large managed to find the best point between the performance (macro F1 = 0.76 ± 0.04 across folds) and the computational efficiency, thus its election as the master model for all coming experiments was justified.

\begin{table}[!ht]
\centering
\caption{Performance of Linear SVC across embedding models on the multilingual unified dataset.}
\label{tab:svc_balanced}
\begin{tabular}{lcccc}
\hline
\textbf{Embedding model}              & \textbf{Accuracy} & \textbf{Precision} & \textbf{Recall} & \textbf{F1} \\
\hline
multilingual-e5-large                 & 0.8018 & 0.7049 & 0.8018 & 0.7330 \\
bge-m3                                & 0.7789 & 0.7002 & 0.7789 & 0.7247 \\
gte-multilingual-base                 & 0.7765 & 0.6869 & 0.7765 & 0.7120 \\
jina-embeddings-v3                    & 0.7795 & 0.6860 & 0.7795 & 0.7103 \\
snowflake-arctic-embed-l-v2.0         & 0.8022 & 0.7085 & 0.8022 & 0.7367 \\
labse                                 & 0.7602 & 0.6661 & 0.7602 & 0.6879 \\
use-multilingual                      & 0.7603 & 0.6800 & 0.7603 & 0.7023 \\
\textbf{xlm-roberta-large}            & 0.7178 & 0.8280 & 0.7178 & 0.7553 \\
\hline
\end{tabular}
\end{table}

\subsection{Inductive Transfer Learning with Data Augmentation (RUN 1)}
To mitigate data scarcity, a defined back-translation augmentation strategy was implemented. Each tweet in its original language was translated to the two alternate languages using the OpenAI GPT-4o-mini API (model="gpt-4o-mini", temperature=0.0 for deterministic output, top\_p=1.0). The translation prompt explicitly requested that semantic content be preserved while acknowledging natural language variation. Each original tweet generated three variants that includes native language, and two back-translated from alternate languages. This one-to-many translation schema tripled the effective training corpus size from N to approximately 3N samples. At the same time the reclamatory class maintaining its original 9-24\% distribution (i.e. as shown in Figure \ref{fig:image2} the class imbalance ratio was preserved during augmentation).

The RoBERTa variant\footnote{https://huggingface.co/cardiffnlp/twitter-xlm-roberta-base, Accessed on January 2026.} taken as a foundation model and finetuned using a custom training loop implementing dynamic undersampling at the epoch level. A custom Batch Sampler class was implemented with the iterative method reconstructed at each epoch to enforce dynamic sampling. For each training epoch, the sampler maintained a 1:3 positive-to-negative ratio by separating dataset indices into positive (label=1) and negative (label=0) groups, drawing exactly one positive sample without replacement for each batch, and drawing exactly three negative samples without replacement. This implementation ensures stochastic variation in negative sample selection across epochs, exposing the model to diverse negative examples while maintaining consistent class ratio. 

\begin{figure}[!ht]
  \centering
  \begin{subfigure}[b]{0.48\textwidth}
    \centering
    \includegraphics[width=\textwidth]{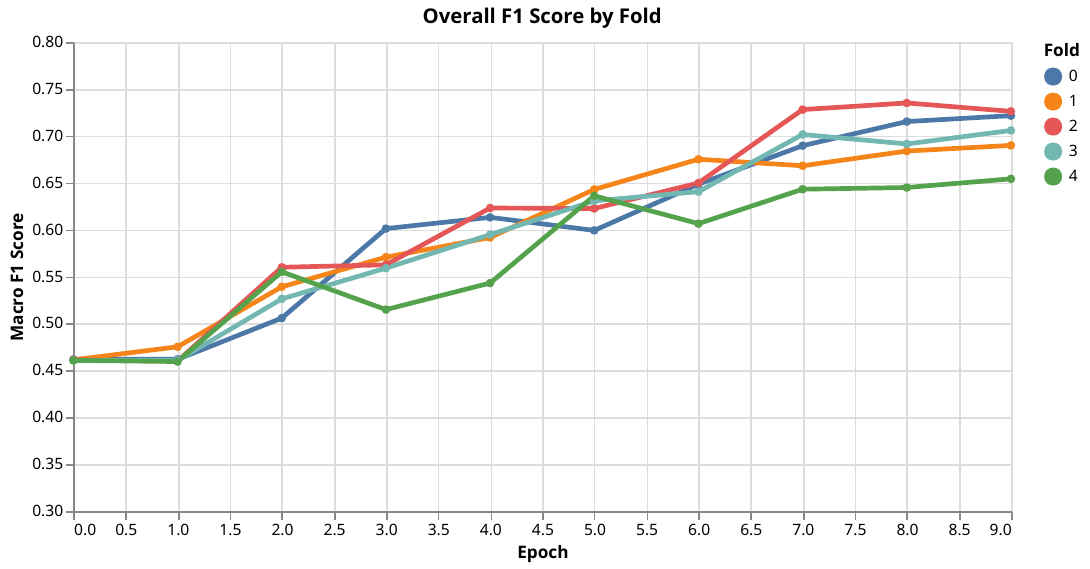}
    \caption{Epoch VS F1 Score}
    \label{fig:image4}
  \end{subfigure}
  \hfill
  \begin{subfigure}[b]{0.48\textwidth}
    \centering
    \includegraphics[width=\textwidth]{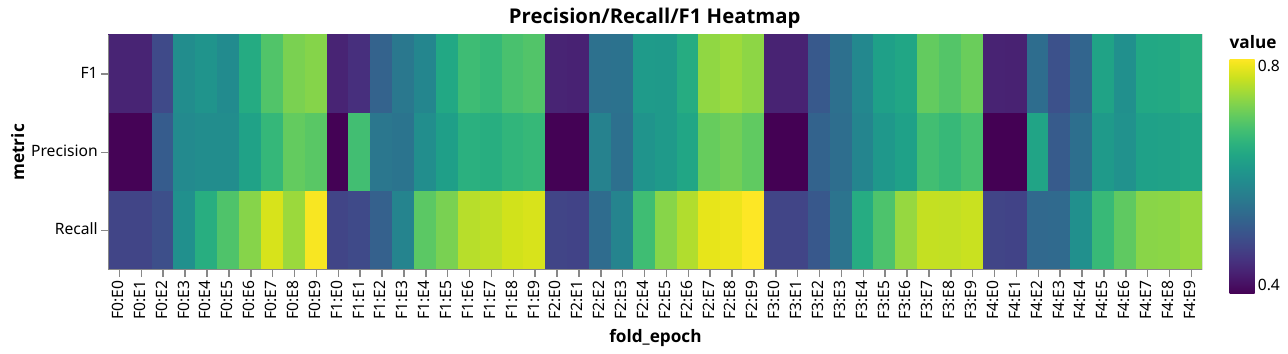}
    \caption{Epoch VS Performance Metrics}
    \label{fig:image5}
  \end{subfigure}
  \caption{Fold Level Performance Metrics for Inductive Transfer Learning.}
  \label{inductive}
\end{figure}

Hyperparameter optimization was performed using Optuna with the Tree structured Parzen Estimator (TPE) as the underlying sampler \cite{akiba2019optuna}. The TPE sampler is still keeping the probabilistic model of the objective function, gradually balancing the step of exploration in the regions of hyperparameters not yet tested with the step of exploitation of the areas around the previously observed good trials that are considered as promising ones. The Optuna objective function encompassed a 5-fold stratified cross validation loop through which the hyperparameter exploration was limited to 50 trials. The hyperparameter search space included: learning rate (log-uniform, 1e-5 to 5e-4), batch size (categorical: {16, 32, 64}), weight decay (log-uniform, 1e-5 to 1e-2), and dropout rate (uniform, 0.1 to 0.4). Early trial termination was implemented via Optuna's Median Pruner with a patience threshold of 3 epochs. The trials whose validation F1 fell below the median of completed trials at the same epoch were automatically pruned. This reduces computational overhead without sacrificing final model quality. Epoch and fold level performance are shown in Figure \ref{inductive}.

Training was conducted for a maximum of 10 epochs per fold using the AdamW optimizer ($\beta_{1}=0.9$, $\beta_{2}=0.999$, $\epsilon=1e-8$) with linear learning rate warmup over the first 10\% of total training steps, followed by linear decay to zero over remaining steps. The loss function was weighted cross entropy, computed as $L = -[w_{0}log(p_{0}) + w_{1} log(p_{1})]$. In which $w_{0}$ and $w_{1}$ are class weights inversely proportional to class frequencies ($w_{0}  \approx  0.35$, $w_{1}  \approx  1.00$ for the imbalanced dataset. Model checkpoints were saved at each epoch based on macro-averaged F1 score on validation set. The best checkpoint for each fold was selected as the model exhibiting the highest validation F1. This configuration produced run 1, representing inductive transfer learning with augmentation and dynamic undersampling.

\subsection{Domain Knowledge Integration via Masked Language Modeling (RUN 2)}
To inject domain specific linguistic knowledge, the foundation model underwent a secondary MLM pretraining phase prior to downstream task finetuning \cite{gururangan2020don}. During MLM, 15\% of tokens in each sequence were randomly selected via Bernoulli sampling and replaced with the [MASK] token. Then the model was tasked with predicting the original tokens from contextual representations. This objective encourages the model to develop deeper representations of LGBTQ+ discourse patterns and reclamatory language usage in multilingual social media contexts.

\begin{figure}[!ht]
  \centering
  \includegraphics[width=\linewidth-80pt]{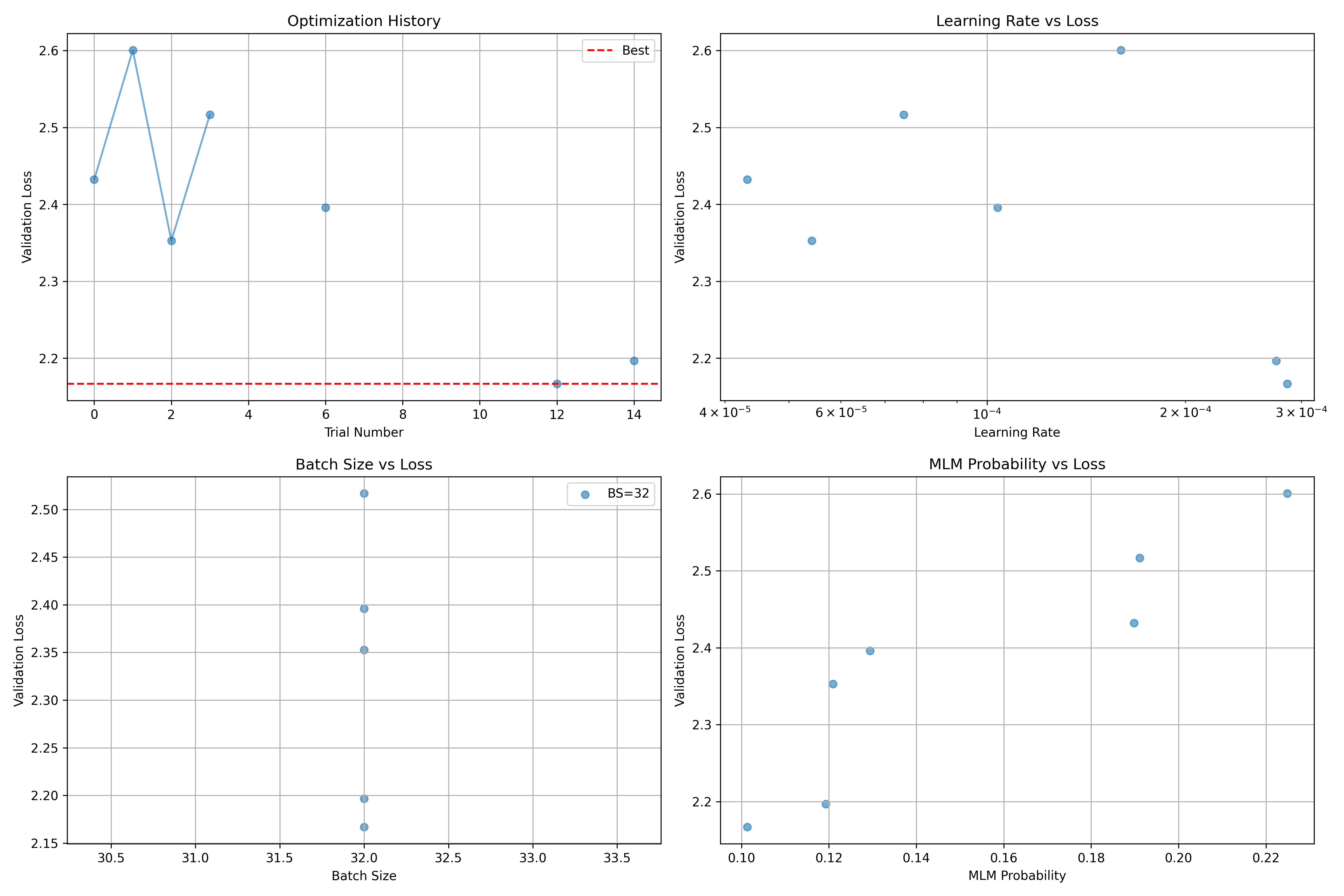}
  \caption{Transductive Transfer Learning: Parameter VS Validation Loss}
  \label{trans}
\end{figure}

The MLM pretraining phase was conducted on the augmented dataset for a maximum of 5 epochs using the AdamW optimizer with hyper parameters identified via a preliminary Optuna search. The MLM search space includeds parameters like learning rate (log-uniform, 1e-5 to 5e-4), batch size (categorical: {16, 32, 64}), weight decay (log-uniform, 1e-5 to 1e-2), and dropout (uniform, 0.1 to 0.4). Optuna evaluated for 50 trials which optimized for minimum validation cross-entropy loss on the MLM task. Pruning was again applied via MedianPruner to terminate unpromising trials early. The MLM loss function was standard cross-entropy $L\_MLM = -\sum log(P\_{\theta}(original\_token|masked\_context))$, which is summed over all masked positions in a batch. The validation loss with respect to parameter are shown in Figure \ref{trans}.

\begin{figure}[!ht]
  \centering
  \begin{subfigure}[b]{0.48\textwidth}
    \centering
    \includegraphics[width=\textwidth]{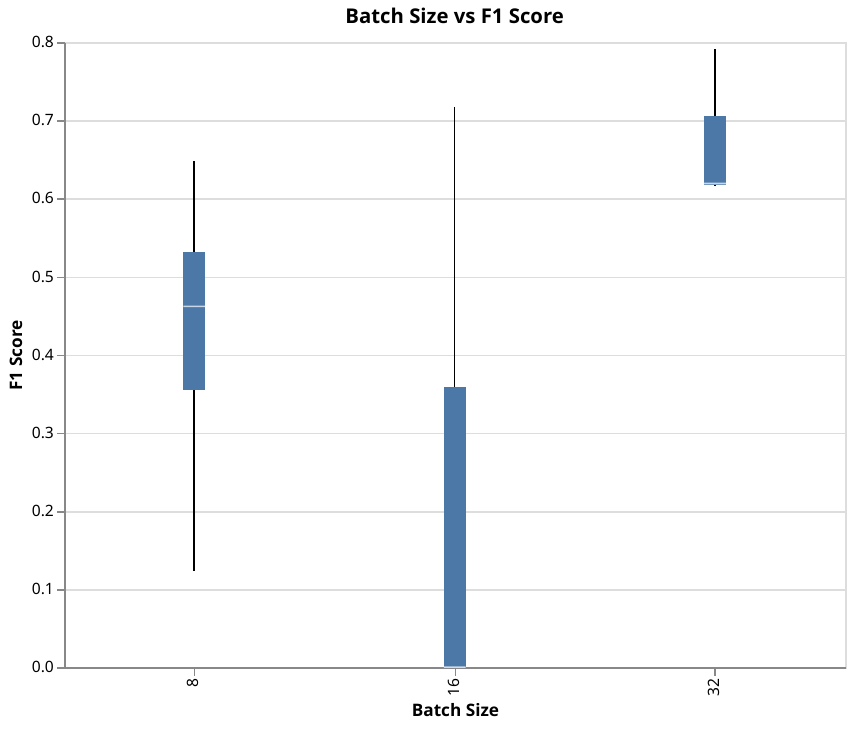}
    \caption{Batch Size VS F1 Score}
    \label{fig:image6}
  \end{subfigure}
  \hfill
  \begin{subfigure}[b]{0.48\textwidth}
    \centering
    \includegraphics[width=\textwidth]{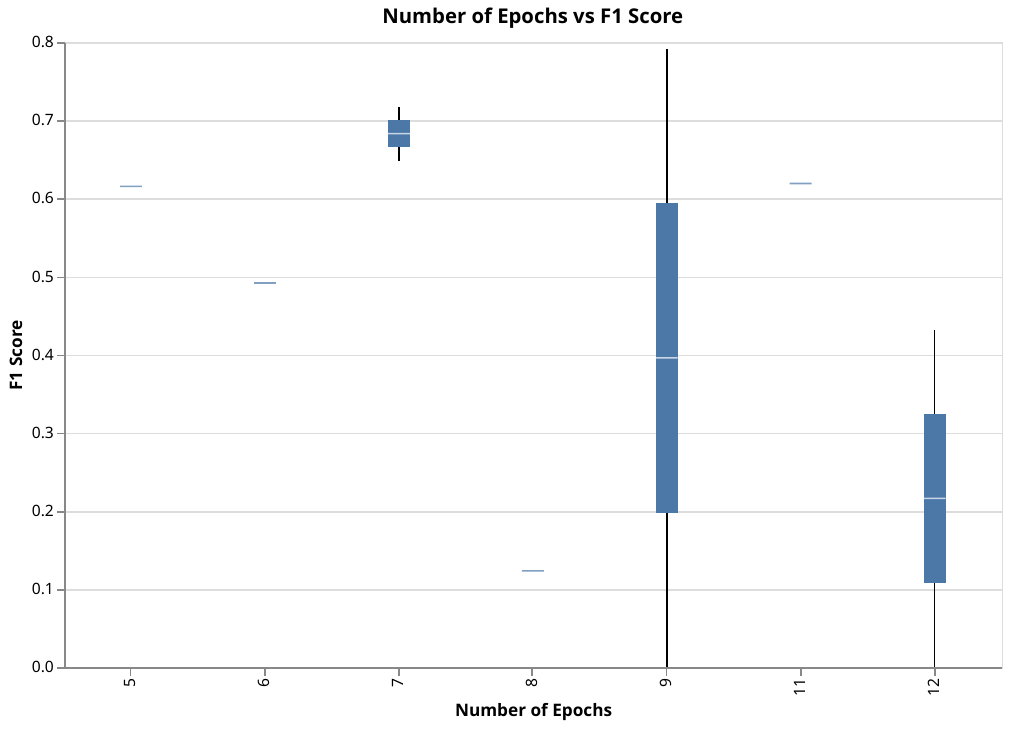}
    \caption{Epoch VS F1 Score}
    \label{fig:image7}
  \end{subfigure}

  \vspace{0.5cm}

 \centering
  \begin{subfigure}[b]{0.48\textwidth}
    \centering
    \includegraphics[width=\textwidth]{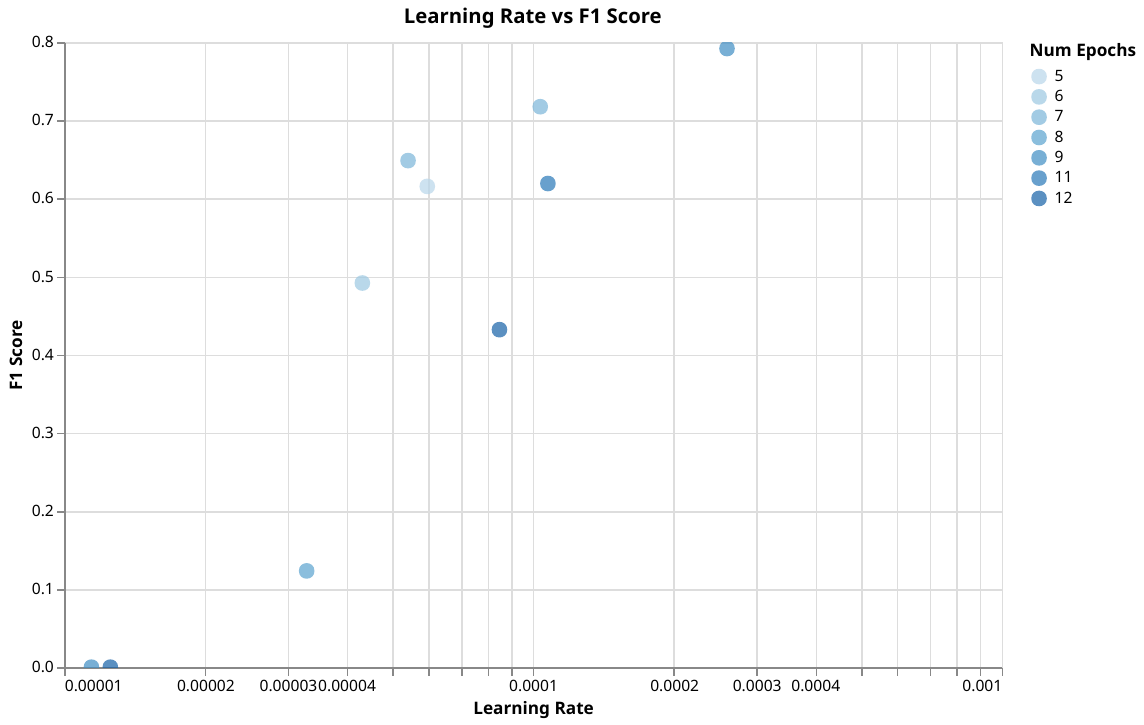}
    \caption{Learning Rate VS F1 Score}
    \label{fig:image8}
  \end{subfigure}
  \hfill
  \begin{subfigure}[b]{0.48\textwidth}
    \centering
    \includegraphics[width=\textwidth]{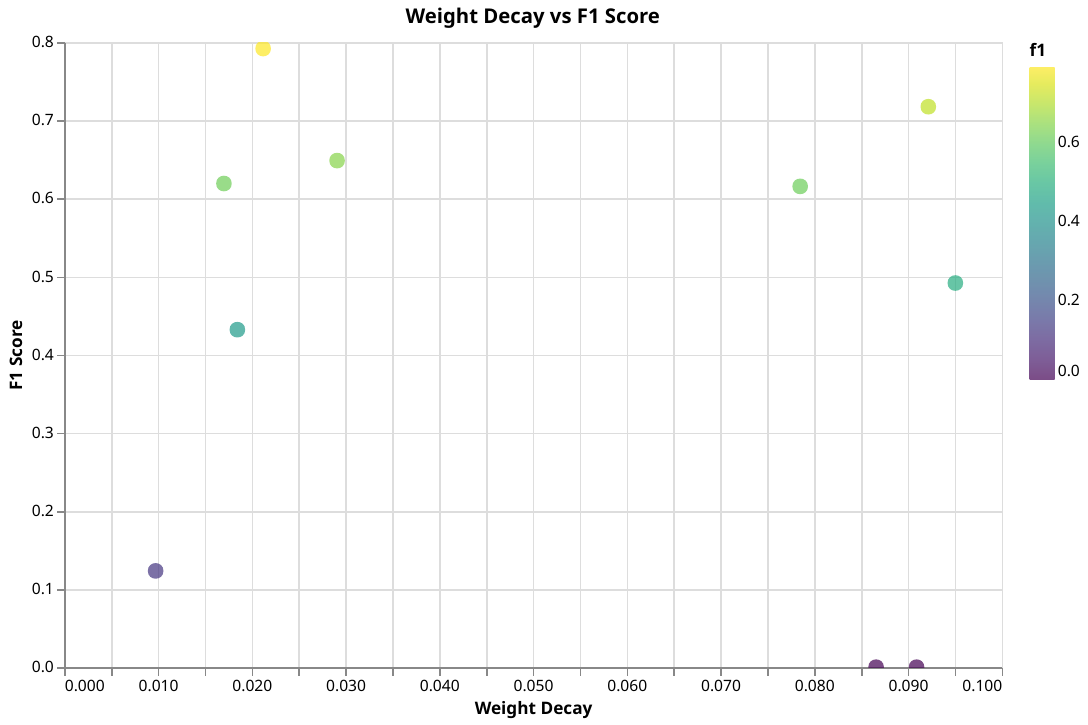}
    \caption{Weight Decay VS F1 Score}
    \label{fig:image9}
  \end{subfigure}

  \caption{Inductive Transfer Learning on Transductive Model. Impact on F1 Score with respect to the parameters.}
  \label{inductive_transductive}
\end{figure}

Following MLM adaptation, the finetuned model was saved and subsequently used as the initialization for downstream finetuning task. This downstream finetuning pipeline was identical to run 1 where dynamic undersampling (1:3 ratio), Optuna hyperparameter optimization (50 trials, TPE sampler, MedianPruner), 5-fold stratified cross validation, and 10 epoch training with early stopping was performed. The same hyperparameter search space as run 1 was employed, allowing direct comparison of the marginal contribution of MLM adaptation. Performance metrics collected after downstream finetuning revealed the cumulative effect of both MLM adaptation and task specific optimization, resulting run 2. Empirical analysis comparing run 1 and run 2 validation F1 scores quantified the absolute and relative improvement attributable to domain knowledge injection via MLM. The impact of inductive transfer learning on transductive model with respect to the hyper parameters are shown in Figure \ref{inductive_transductive}.

\subsection{Language-Specific Threshold Refinement and Prediction Reclassification}
Both run 1 and run 2 models produced continuous confidence scores via softmax normalization of the final layer logits $conf\_score = exp(logit\_1) / (exp(logit\_0) + exp(logit\_1))$, where $logit\_0$ and $logit\_1$ denote the class-specific logits. Default binary classification employs a threshold of 0.5 with equal misclassification costs and balanced class priors. In the MultiPRIDE task, neither assumption holds where the dataset exhibits class imbalance (9-24\% minority class), and language-specific label distributions differ significantly ($ \chi^2$ p < 0.001).

To optimize decision boundaries, Receiver Operating Characteristic (ROC) curve analysis was performed independently for each language. For each language (English, Spanish, Italian), validation set confidence scores were analyzed across a fine-grained threshold range. For each candidate threshold $\tau$, performance metrics were computed includes true positive rate $(TPR) = TP/(TP+FN)$, false positive rate $(FPR) = FP/(FP+TN)$, and $F1 score = 2·TP/(2·TP+FP+FN)$. The final selection for the optimal threshold $\tau*$ of each language was based on the value that provided the maximum macro-averaged F1 score on the validation set, thus safeguarding equal performance of both the minority and the majority classes through their simultaneous consideration.

\begin{figure}[!ht]
  \centering
  \includegraphics[width=\linewidth-80pt]{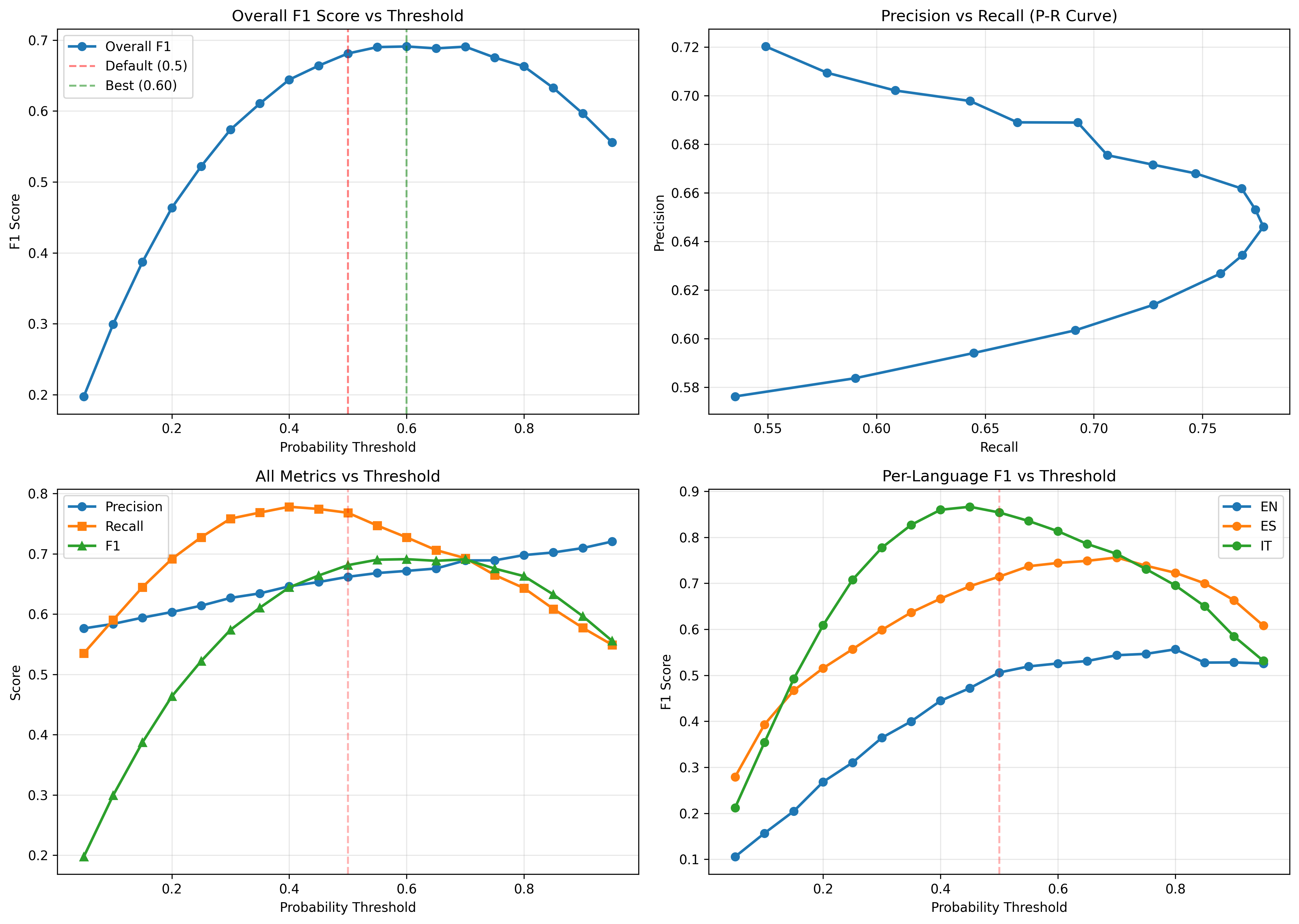}
  \caption{Threshold Analysis: Language-specific Optimal Thresholds}
  \label{threshold}
\end{figure}

It is interesting to point out the variation of the optimal thresholds being language dependent, thus, the differences in model confidence predictions were revealed. For instance, it was found that English tweets had quite high average confidence scores for the reclamatory class which resulted in the necessity of the higher threshold (say, $\tau\_{en} = 0.58$) to ensure the restriction of precision. On the other hand, Italian tweets had lower average confidence necessitating a lower threshold (for instance, $\tau\_{it} = 0.42$) to capture the instances of the minority class. Spanish thresholds were usually around these extremes ($\tau\_{es}  \approx  0.50$). These thresholds were subsequently enforced on the predictions so that if $conf\_score \geq  \tau\_{language}$, the prediction will be reclamatory, otherwise, it will be non-reclamatory. The per language threshold variation has been represented in Figure \ref{threshold}.

% The optimal threshold $\tau*$ for each language was selected finally as the value maximizing macro-averaged F1 on the validation set, ensuring equal weighting of minority and majority class performance.

% Notably, optimal thresholds exhibited language specific variation, reflecting distributional differences in model confidence predictions. For an example, analysis revealed that English tweets exhibited relatively high average confidence scores for the reclamatory class, necessitating a higher threshold (e.g., $\tau\_{en} = 0.58$) to maintain precision. Conversely, Italian tweets showed lower average confidence, requiring a lower threshold (e.g., $\tau\_{it} = 0.42$) to capture minority class instances. Spanish thresholds typically fell between these extremes ($\tau\_{es}  \approx  0.50$). These thresholds were applied post prediction via setting prediction = reclamatory if $conf\_score \geq  \tau\_{language}$ else non-reclamatory. This threshold variation per language is highlighted in Figure \ref{threshold}. 

% The run 1 model predictions were reclassified using learned language specific thresholds, yielding run 3. Similarly, run 2 predictions were reclassified, yielding run 4. This threshold refinement represents a critical post prediction optimization step that does not require additional computational requirements. This step typically yields 2-5\% absolute F1 improvement by adapting decision boundaries to empirically observed confidence distributions.

The predictions of run 1 model were reclassified by means of learned language specific thresholds giving rise to run 3. In the same way, reclassification of run 2 predictions was done yielding run 4. The refining of the thresholds is a very crucial post-prediction optimization step which does not require extra computational power. This step usually results in a 2-5\% absolute F1 improvement by the adaptation of decision boundaries to the empirically observed confidence distributions.

\begin{figure}[!ht]
  \centering
  \includegraphics[width=\linewidth-80pt]{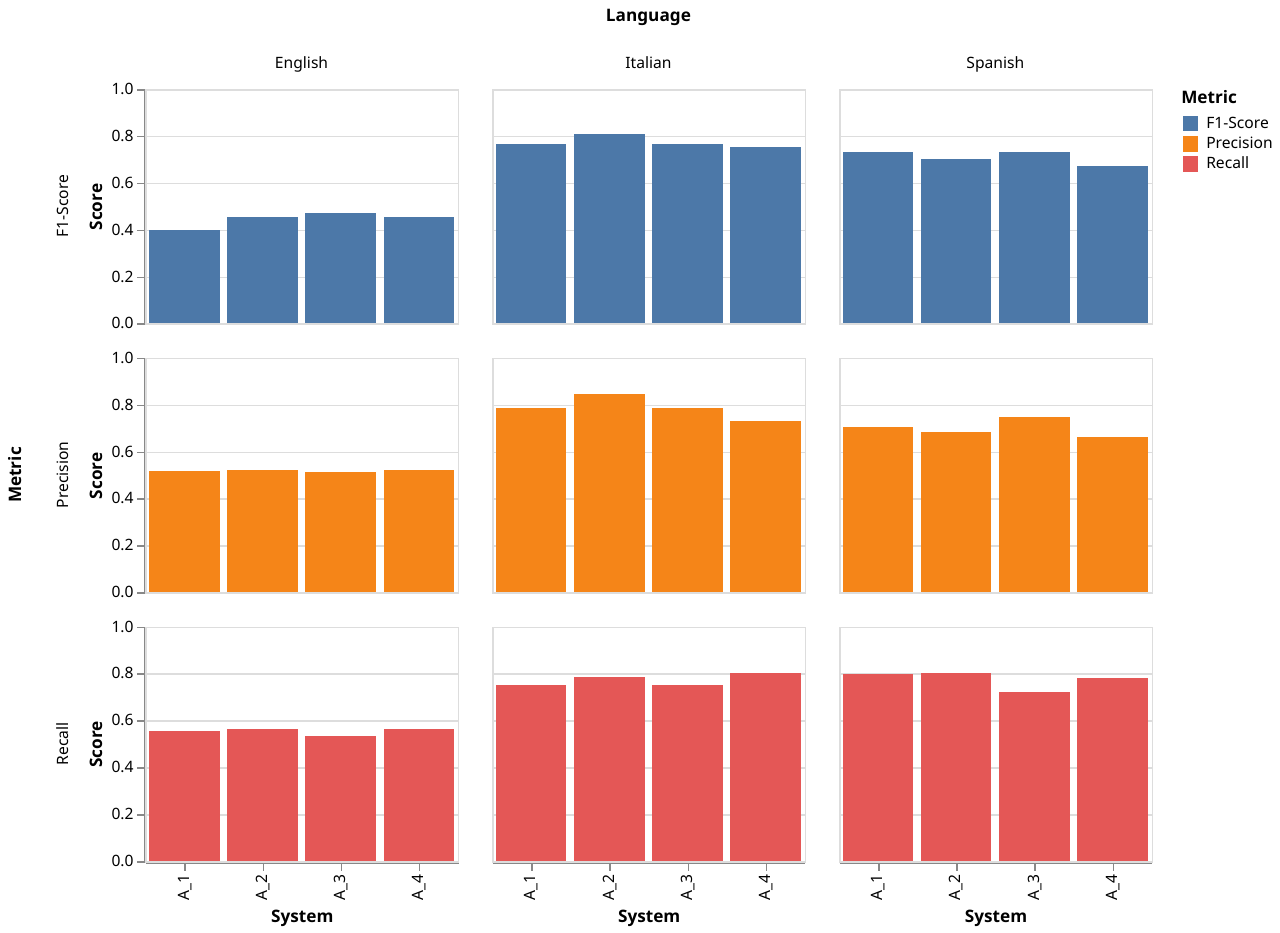}
  \caption{Final Test Set Results of Submitted Runs 1 - 4.}
  \label{test_results}
\end{figure}

\section{Observations}
\label{sec:observations}

% The final submitted results on test set is shown in Figure \ref{test_results}. Based on the experimental findings across all the runs, the framework demonstrates that principled treatment of data scarcity, class imbalance, and multilingual variation yields significant performance gains in reclamation classification task. 

Figure \ref{test_results} presents the final submitted results on the test set. The overall results from all the runs revealed the framework's ability to deliver considerable performance improvements in the reclamation classification task by the proper management of data scarcity, class imbalance, and multilingual variation. The foundation model selection outcomes at Table \ref{tab:svc_balanced} revealed that RoBERTa pretrained on social media context obtains superior macro-averaged F1 stability (0.7553 ± 0.04). The recall value is also stable which is essential for minimizing missed reclamatory instances in low resource contexts. This selection decision prioritizes minority class detection over overall accuracy. This is considered where false negatives carry higher sociolinguistic cost than false positives in LGBTQ+ discourse analysis. The back translation augmentation strategy successfully tripled the effective training corpus while preserving language specific distributional characteristics. This preservation of label ratio proved non-trivial with the dynamic epoch level under sampling at a 1:3 positive-to-negative ratio. It is implemented with stochastic negative sample selection which prevented both majority class collapse and overfitting to repeated examples. The combination of weighted cross entropy loss ($w_{1}$ $\approx$ 1.00 for minority class) and consistent sampling constrain created stable convergence visible in fold wise F1 stabilization by epoch 6–7. This demonstrates that class balance can be achieved without aggressive resampling side effects.

The integration of domain knowledge through MLM (RUN 2) therefore gave language-dependent results, showing that MLM adaptation is not universally advantageous across multilingual contexts. While English performance showed a marginal improvement, Spanish and Italian displayed more variable responses to MLM pre-training. It would appear that morphologically rich languages and limited pre-training data create different optimization landscapes. The hyperparameter sensitivity analysis in Figure \ref{inductive_transductive} shows that MLM adapted models require more precise tuning: batch size, learning rate and weight decay all display non-monotonic relationships to F1 scores. While optimal configurations residing in narrow ranges, specifically: batch size 16–32, learning rate 1e-4 to 3e-4, and weight decay 0.08–0.09. This underlines the importance of careful hyperparameter selection for domain adaptation. The Optuna TPE sampler evidenced efficient convergence by trial 30–35 within our 50-trial budget, confirming the computational practicality of the system for shared task participation by avoiding exhaustive grid search.

The most interpretable findings of Figure \ref{threshold} reveal the threshold optimization that is specific to different languages. The decision boundaries that are chosen for reclamation detection are actually dependent on the languages: for English, it is 0.58, for Spanish, it is 0.50, and for Italian, it is 0.42. The range of difference is 16\% and it is due to the diversity of languages in the context of culture-based reclamation and to the distribution of model confidence scores that are determined by the characteristics of training data. Consequently, English tweets that have more consistent reclamation markers produce higher average confidence scores, and thus, require higher thresholds to keep the precision. On the other hand, the Italian tweets that have dependent and more subtle cues in the context require lower thresholds to find the true positives. The significance in computation and the absolute F1 gains that are noticeable through this threshold refinement point out that a 0.5 default threshold assumption is a violation of linguistic universality. To conclude, by analyzing the per-language test sets, the asymmetries of the generalization patterns illustrated in the figure are revealed. The languages that have the highest native reclamatory prevalence (Italian 24\%) exhibit different precision-recall dynamics compared to the low-prevalence languages such as English at 9\%, where recall improvements are disproportionately benefiting the low-prevalence cases where class imbalance is most pronounced.

The scrutinization of the 60 misclassified samples and the final test results depicted in Figure \ref{test_results} not only highlights the linguistic and English-specific ambiguities but also hints at the social aspect of language and its complexity in daily communication as the problem. The English category has mistakes mostly in the areas where irony, hypothetical discourse, and theoretical discussion seem to intermingle with the reclamation of the terms used for the stigma. Examples like "Does this mean I can start calling you a faggot?" (sarcasm questioning permission) and meta-textual discourse on the reclamation terminology (e.g., debating whether "faggot" could be a form of self-affirmation in different situations) are correctly classified as non-reclamatory but yet, the system misclassifies them as reclamatory. The problem is with the model as it cannot draw a line between the two uses that are covering "I'm gay and you're a faggot" which is said in a sarcastic way and the real reclamation. This is indeed what calls for knowing the speaker's intent and the social context which is more than just the lexical word features. The Spanish language mistakes reveal different failure modes of the system like its getting stuck in the race-reclaiming in activist contexts where slurs are actually reclaimed just to take a stand against the heteronormative structures (e.g. "orgullosísimo de ser un peazo maricón" representing pride in activist framing). The Italian errors, in contrast, are mainly negative false where real reclamatory instances go unnoticed. Such cases contain the subtle cultural markers of reclamation where the intimacy shown through slur usage among in-group members is expressed (e.g. "sono pure ricchione. Bella sis" as solidarity acknowledgment). Likewise, the humor and community bonding through slur appropriation (e.g. "i viaggi ricchioni" as casual community reference) and self-descriptive pride assertions (e.g. "noi ricchione lavoratrici") that lack the explicit affirmative markers the model learned to recognize.

The Italian false negatives reveal a strong correlation with low confidence scores (0.42) and this means the model is not confident in its ability to detect Italian reclamatory patterns. These language specific error modes highlight that reclamation detection in multilingual situations cannot be based on uniform lexical or statistical patterns. English requires modeling of pragmatic context in order to filter out sarcasm and hypothetical discourse. the Spanish case benefits from activist discourse framing. In the Italian case, there is a need for explicit cultural knowledge about in-group solidarity and use of intimate language. The current limitation of only text input increases the performance ceiling especially for the Italian language where non-linguistic cues (tone of voice, emoji sentiment, social relationship signals) would help recognize reclamatory intention. The system using language-specific thresholds still cannot resolve these foundational representational constraints. Such constraints include enriching training data with marking annotations that differentiate sarcasm from reclamation in English, embedding explicit discourse context modeling for activist and political uses in Spanish, and augmenting training with discourse patterns of intimate communities for Italian. The errors are consistent and systematic, the first grouping being according to the languages used and the second according to the pragmatic functions. This presumes that for multilingual reclamation detection to be accurate, not only the use of language aware models but also the understanding of reclamation through a pragmatically informed and culturally grounded approach as different in linguistic communities is a necessity.

\section{Conclusions}
\label{sec:conclusion}

This paper put forward a multistep method for slur reclamation detection to be applied in different linguistic environments. An effective training pipeline for the MultiPride shared task was constructed by using dynamic sampling to address the class imbalance problem and by increasing the dataset with back translation techniques. Testing has shown that the XLM-RoBERTa is a highly effective foundation model but its performance varies a lot between languages, and the main reason is the cultural factor.

One significant finding is that a “universal” decision boundary is not appropriate for the multilingual sentiment analysis. Adjusting the classification thresholds for each language resulted in a 2-5\% improvement in F1 scores without incurring any extra computational costs. This suggests that the models do not have the same certainty levels across languages, with English requiring stricter thresholds to prevent false positives and Italian needing lower thresholds to detect subtle reclamation cases. The framework demonstrates that systematic data imbalance treatments, cross-lingual variation recognition, and domain-specific knowledge application can produce robust multilingual sentiment classifiers that are socially sensitive and applicable to such areas.

Nonetheless, the error analysis shows that the models relying solely on text have trouble grasping the pragmatic nuances. The system often confuses sarcasm with reclamation,especially in English, and fails to recognize implicit solidarity in Italian. The application of domain adaptation via MLM only resulted in limited gains, necessitated intricate hyper-parameter tuning, and more data. Future research should focus on features beyond text and incorporate signals such as emojis or user interaction graphs to better capture the social intent behind the language.

\section*{Declaration on Generative AI}
  
During the preparation of this work, the authors used Generative AI in order to: Grammar and spelling check. After using these tools/services, the authors reviewed and edited the content as needed and take full responsibility for the publication’s content. 

\bibliography{sample-ceur}

\end{document}